\documentclass{svproc}
\usepackage{amsfonts}

\usepackage{mathrsfs}                
\usepackage{bm}                      
\usepackage{amsmath}
\usepackage{verbatim}
\usepackage{graphicx}
\usepackage{url}
\usepackage{float}
\usepackage{subfigure}

\spnewtheorem{assumption}{Assumption}{\itshape}{\rmfamily}

\begin{document}
\mainmatter              
\title{Distributed Safe Cooperative Vector Field for Trajectory Curvature Constrained Multi-Robot Systems}
\titlerunning{Hamiltonian Mechanics}  
%
\author{
    Zhouru Xiao\inst{1}  \and
    Tao Teng\inst{1} \and
    Weijia Yao\inst{1} \and
    Hui Zhang\inst{1} \and
    Xiangke Wang\inst{2} \and
    Yaonan Wang\inst{1}
}
\renewcommand{\thefootnote}{}
\footnotetext{This work was supported in part by the National Natural Science Foundation of China under Grant 62573182, and in part by the State Key Laboratory of Digital-Intelligent Modeling and Simulation.}
\footnotetext{Weijia Yao is the corresponding author.}
\footnotetext{Zhouru Xiao and Tao Teng contributed equally to this work.}
\authorrunning{Zhouru Xiao et al.} 
\tocauthor{Zhouru Xiao, Tao Teng, Weijia Yao, Hui Zhang, Xiangke Wang and Yaonan Wang} 

\institute{
    School of Artificial Intelligence and Robotics, Hunan University, Changsha 410082, China
    \and
    College of Intelligence Science and Technology, National University of Defense Technology,  Changsha 410082, China
}

\maketitle              

\begin{abstract}
Trajectory curvature constraints are inherent in practical multi-robot systems due to the limited turning capabilities of the robots. Without properly accounting for these constraints, robots may fail to accomplish assigned tasks, and their trajectories may diverge from the intended paths. This paper proposes a distributed safe cooperative vector field approach for multi-robot systems subject to trajectory curvature constraints. The proposed approach is composed of a cooperative vector field and a safety-oriented collision avoidance vector field, aiming to address the problems of cooperative motion and safe collision avoidance in multi-robot path-following tasks. A safety-oriented collision avoidance vector field with adaptively adjustable reactive boundary is developed to accommodate the kinematic curvature constraints of robots, thereby ensuring the physical feasibility of collision avoidance maneuvers. The proposed vector field requires only a single virtual variable from each neighboring robot to achieve cooperative motion and ensure both obstacle avoidance and inter-robot collision avoidance. The effectiveness of the proposed approach is validated through both simulations and real-world experiments on an actual multi-robot platform.
\end{abstract}

\section{Introduction}

Distributed cooperative control is an important research area for enabling multi-robot systems (MRS) to efficiently perform tasks in complex environments, including ocean exploration \cite{cao:yu:ren:chen:13}, smart agriculture \cite{A:deep:reinforcement}, and disaster response \cite{Multi:agent:systems}. To achieve multi-robot cooperative control,  a variety of distributed control strategies have been developed. The comprehensive surveys \cite{Anoverview,wang2016multi} provide an overview of the state-of-the-art and emerging trends in distributed cooperative control for multi-robot systems, encompassing consensus, distributed formation control, distributed optimization, distributed task allocation, and distributed estimation and control. Among these methods, the Distributed Vector Field (DVF) approach \cite{yao:mar:sun:cao:23} has attracted considerable attention due to its scalability, inherent distributed computation capability and strong resilience to communication interruptions. The DVF method achieves coordinated motion by treating the path parameter as an additional virtual coordinate within the guiding vector field and sharing this virtual coordinate among neighboring robots.\vspace{-0.1em}

Despite notable advances, the application of Distributed Vector Field (DVF) approaches to heterogeneous multi-robot systems, including ground vehicles, fixed-wing UAVs, and other platforms, reveals limitations that are often overlooked. In particular, existing methods\cite{goncalves2010vector,rezende2022constructive} rarely account for the intrinsic kinematic constraints of individual robots, such as minimum turning radii. Conventional vector field designs primarily aim to achieve cooperative objectives, such as formation maintenance, coverage, or obstacle avoidance, while often assuming idealized robots with unconstrained mobility, for example, omnidirectional motion. In practice, robotic platforms are subject to physical and dynamic limitations and cannot instantaneously change their motion direction. When the rate of change of the vector field’s guidance exceeds a robot’s maximum curvature, the resulting trajectories may be dynamically infeasible because the robot cannot follow the commanded direction and speed within its operational limits. Such infeasibility can induce tracking errors, compromise system stability, and increase the risk of inter-robot and robot-obstacle collisions, particularly in dense or tightly coordinated scenarios. Although curvature-constrained path planning and trajectory tracking for single robots have been extensively studied \cite{lau:ede:oet:15,shi:rat:21,she:wil:gup:19}, systematically and distributively integrating these constraints into multi-robot cooperative vector field design, while ensuring dynamic feasibility under strict collision and obstacle avoidance requirements, remains a significant open challenge.

To address the challenges discussed above, this paper proposes a  distributed safe cooperative control algorithm for trajectory curvature-constrained heterogeneous multi-robot systems. Building upon previous work on distributed cooperative vector fields \cite{yao:mar:sun:cao:23} and composite collision avoidance vector fields \cite{yao:lin:and:cao:22}, we propose a novel Distributed Safe Cooperative Vector Field (DSCVF), specifically designed for heterogeneous multi-robot systems operating in planar environments under curvature constraints. The central contribution of this work is the explicit integration of kinematic curvature constraints into the vector field design, ensuring that each robot’s motion control inputs during cooperative maneuvers remain within its physical maneuvering capabilities (i.e., respecting the minimum turning radius) and can be stably tracked. Additional contributions of this paper include:

1) Design of a safety-oriented collision avoidance vector field with an adaptively adjustable reactive boundary. The level set values of the vector field’s reactive boundary can be adaptively adjusted online based on each robot’s pose and the location of neighboring obstacles, enabling reliable and efficient collision avoidance.

2) Construction of a distributed safe cooperative vector field for curvature-constrained heterogeneous robots. By composing cooperative vector fields with the safety-oriented collision avoidance vector field, the proposed DSCVF retains the scalability and low communication load of classical DVF approaches while offering potential extensions to more complex scenarios, such as motion planning in higher-dimensional spaces.

\textbf{Notations:} Let the integer set be denoted by $\mathbb{Z}_i^j := \{ m \in \mathbb{Z} : i \le m \le j \}$, where $i, j \in \mathbb{Z}$ and $i \leq j$. Boldface letters denote vectors $\mathbf{v} \in \mathbb{R}^n$ in the $n$-dimensional Euclidean space, with the $i$-th component denoted as $v_i$, where $i \in \mathbb{Z}_1^n$. \(\hat{(\cdot)}\) denotes the normalization operator, i.e., for any nonzero vector \(v \in \mathbb{R}^n\), \(\hat{v} = \frac{v}{\|v\|}\). The distance between a point \(p_0\) and a nonempty set \(\mathcal{S}\) is defined as $\operatorname{dist}(p_0, \mathcal{S}) = \inf \{ d(p, p_0) : p \in \mathcal{S} \}$, where \(d(p, p_0) = \|p - p_0\|_2\) denotes the Euclidean distance.

\textbf{Graphs:} The set of nodes $\mathcal{V} := \{1, \ldots, n\}$ represents the set of robots, and the edge set $\mathcal{E} \subseteq (\mathcal{V} \times \mathcal{V})$ encodes the communication links. The neighbor set of robot $i$ is defined as $\mathcal{N}_{i} := \{ j \in \mathcal{V} : (i, j) \in \mathcal{E} \}$. This paper considers only undirected graphs; that is, if $(i, j) \in \mathcal{E}$, robots $i$ and $j$ can exchange information bidirectionally. For an introduction to graph theory, see \cite{MesbahiEgerstedt+2010}.

\section{Problem Formulation}\label{er}
Consider \(n\) robots with unicycle kinematics operating in an environment \(W\). These robots communicate in a distributed manner over a graph \(\mathcal{G} = (\mathcal{V}, \mathcal{E})\). The environment contains \(m\) static obstacles. The kinematic model of the \(i\)-th robot for $i \in \mathbb{Z}_1^n$ is given by:\vspace{-0.5em}
\begin{equation}\label{0001}
\dot{x}_i = v_i \cos\theta_i,\quad 
\dot{y}_i = v_i \sin\theta_i,\quad 
\dot{\theta}_i = u_i,
\end{equation}
where \(\boldsymbol{\zeta}_i = [\boldsymbol{\xi}_i, \theta_i]^\top \in \mathbb{R}^2 \times \mathbb{S}^1\) denotes the configuration vector of the $i$-th robot, consisting of the position \(\boldsymbol{\xi}_i = [x_i, y_i]^\top \in \mathbb{R}^2\) and the heading angle \(\theta_i \in \mathbb{S}^1\) with respect to the global Cartesian coordinate frame. The linear velocity \(v_i\) is constant, and the angular velocity \(u_i \in [-\bar{u}_i,\,\bar{u}_i]\) serves as the control input. The curvature of the trajectory of the \(i\)-th robot  is
$\kappa_i = \left|\frac{u_i}{v_i}\right| \le \bar{\kappa}_i$,
where \(\bar{\kappa}_i\) is the maximum allowable curvature for the \(i\)-th robot. Accordingly, the minimum turning radius of the \(i\)-th robot is defined as\vspace{-0.5em}
\begin{equation}\label{1212}
\rho_i = \frac{1}{\bar{\kappa}_i} = \frac{v_i}{\bar{u}_i}. 
\end{equation}
Therefore, since the angular velocity constraints and maximal allowable curvature constraints are different, the multi-robot system is \emph{heterogeneous} in this sense. 

Each robot is modeled as a closed disk with radius \(r_c\). Each static obstacle \(\mathcal{O}_i\) for \(i \in \mathbb{Z}_1^m\) is represented by a closed disk centered at \(\boldsymbol{\xi}_{i}^o = [x_{oi}, y_{oi}]^\top\) with radius \(r_i\), i.e., $\mathcal{O}_i := \left\{ \boldsymbol{\xi} \in \mathbb{R}^2 \mid \varphi_i=\|\boldsymbol{\xi} - \boldsymbol{\xi}_{i}^o\|_2^2 - r_i^2=0 \right\}$, where \(\varphi_j: \mathbb{R}^2 \to \mathbb{R}\) is  a twice continuously differentiable function. The $i$-th robot is required to follow a path $\mathcal{P}_i$ in $\mathbb{R}^2$, parameterized by two scalar functions $x_{i1} = f_{i1}(w_i)$ and $x_{i2} = f_{i2}(w_i)$, where $x_{ij}$ denotes the $j$-th coordinate, $f_{ij} \in C^2$ is the $j$-th parametric function of the $i$-th robot and  $w_i \in \mathbb{R}$ is the path parameter, for $i \in \mathbb{Z}_1^n, j \in \mathbb{Z}_1^2$. Accordingly, the desired path (in a higher dimensional space)
 can be expressed as
$\mathcal{P}_i := \{ \bar{\boldsymbol{\xi}}_i \in \mathbb{R}^3 \mid \phi_{ij}(\boldsymbol{\xi}) = 0, \, j \in \mathbb{Z}_1^2 \}$, where $\bar{\boldsymbol{\xi}}_i = (x_{i1}, x_{i2}, w_i) \in \mathbb{R}^3$ represents the \emph{generalized coordinate}, and the surface functions are defined by $\phi_{ij} := x_{ij} - f_{ij}(w_i)$ for $j \in \mathbb{Z}_1^2$\cite{yao:mar:lin:cao:21}. We require two mild standing assumptions that are reasonable in practical applications:
\begin{assumption}
The communication graph \(\mathcal{G} = (\mathcal{V}, \mathcal{E})\) is connected.
\end{assumption}
\begin{assumption}
For each $i \neq j \in \mathbb{Z}_1^m$, dist($\boldsymbol{\xi}_{i}^o,\boldsymbol{\xi}_{j}^o)>r_i+r_j+2r_c$.
\end{assumption}

Given the desired paths $\mathcal{P}_i$, starting from a specific reference configuration $\boldsymbol{w}^* = (w_1^*, \ldots, w_n^*)^\top$, the desired relative states $\Delta_{ij}$ are constructed by calculating the difference between relevant reference variables. Namely, the stacked vector $\boldsymbol{\Delta}^* = D^\top \boldsymbol{w}^*$ collects all $\Delta_{ij}$ for edges $(i,j) \in \mathcal{E}$, where $D \in \mathbb{R}^{n \times |\mathcal{E}|}$ is the incidence matrix of the graph, which encodes the communication topology among robots. The primary objective is to design the angular control input $\boldsymbol{u}=[u_1,u_2,\dots,u_n]^\top$ in (\ref{0001}), subject to the minimum turning radius constraint (\ref{1212}), such that cooperative motion, obstacle avoidance and inter-robot collision avoidance can be achieved. Namely:

1) \textbf{Cooperative motion}: In the absence of obstacles, for any $(i, j) \in \mathcal{E}$, it holds that $\lim_{t \to \infty} (\omega_i(t)-\omega_j(t)-\Delta_{ij})=0$.

2) \textbf{Obstacle avoidance}: For any \(i \in \mathbb{Z}_1^n\), \(j \in \mathbb{Z}_1^m\), and \(t > 0\), it holds that $\|\boldsymbol{\xi}_i(t) - \boldsymbol{\xi}_{j}^o\| > r_c + r_j$.

3) \textbf{Inter-robot collision avoidance}: For any distinct \(i, j \in \mathbb{Z}_1^n\), it holds that $\|\boldsymbol{\xi}_i(t) - \boldsymbol{\xi}_j(t)\| > 2 r_c$.

\section{Design of Safe Cooperative Vector Field}\label{san}

\subsection{Distributed Cooperative Vector Field}
Distributed cooperative vector fields were initially proposed in \cite{yao:mar:sun:cao:icra:21} to achieve path following and motion coordination of multiple robots on general one-dimensional manifolds. According to \cite{yao:mar:lin:cao:21}, the high-dimensional guiding vector field for the $i$-th robot, $\mathcal{X}_i : \mathbb{R}^3 \to \mathbb{R}^3$, is designed as follows:\vspace{-0.5em}
\begin{equation}
\label{0003}
\mathcal{X}_i(\bar{\boldsymbol{\xi}}_i) = \times \big(\nabla \phi_{i1}, \nabla \phi_{i2}\big) - \sum_{j=1}^2 k_j \phi_{ij} \nabla \phi_{ij},
\end{equation}
where $k_{ij} > 0$, $\nabla \phi_{ij} \in \mathbb{R}^3$ denotes the gradient of the function $\phi_{ij}$ with respect to the generalized coordinate $\bar{\boldsymbol{\xi}}_i$, and $\times(\cdot)$ denotes the vector cross product. The physical meaning of this vector field is as follows: the first term $\times(\nabla \phi_{i1}, \nabla \phi_{i2})$ is orthogonal to all gradient vectors and provides a tangential propagation direction along the desired path; the second term $-\sum_{j=1}^2 k_j \phi_{ij} \nabla \phi_{ij}$ is a weighted sum of the gradients used to attract the robot trajectory toward the intersection of the surfaces, i.e., the desired path. Consequently, the explicit form of the path-following guiding vector field for the $i$-th robot, $^{\mathbf{pf}}\boldsymbol{\mathcal{X}_i} : \mathbb{R}^3 \to \mathbb{R}^3$, is given by:
\begin{equation}\label{0004}
^{\mathbf{pf}}\boldsymbol{\mathcal{X}_i}(\bar{\boldsymbol{\xi}}_i) =
\begin{bmatrix}
f_{i1}^{\prime}(w_i) - k_{i1} \phi_{i1}(\bar{\boldsymbol{\xi}}_i) \\
f_{i2}^{\prime}(w_i) - k_{i2} \phi_{i2}(\bar{\boldsymbol{\xi}}_i) \\
1 + \sum_{j=1}^2 k_{ij} \phi_{ij}(\bar{\boldsymbol{\xi}}_i) f_{ij}^{\prime}(w_i)
\end{bmatrix},
\end{equation}
where $f_{ij}^\prime$ denotes the derivative of $f_{ij}$ with respect to its argument $w_i$.

To achieve coordination of $w_i$, thereby indirectly realizing position coordination among multiple robots, a coordination component for the $i$-th robot, $^{\mathbf{co}}\boldsymbol{\mathcal{X}_i} : \mathbb{R}^n \to \mathbb{R}^3$, is introduced as:
\begin{equation}\label{0005}
^{\mathbf{co}}\boldsymbol{\mathcal{X}_i}(\boldsymbol{w}) = \begin{pmatrix}0, 0, c_i(\boldsymbol{w}) \end{pmatrix}^\top,
\end{equation}
where $\boldsymbol{w} = (w_1, \ldots, w_n)^\top$, and $c_i : \mathbb{R}^n \to \mathbb{R}$ is a coordination function enabling local interaction through neighboring virtual coordinates $w_j$ ($j \in \mathcal{N}_i$). Specifically, the difference $w_i(t) - w_j(t)$ is driven to converge to the desired relative state $\Delta_{ij} \in \mathbb{R}$ for each edge $(i,j) \in \mathcal{E}$. The design of $c_i$ is detailed below. 

First, we adopt the following consensus control algorithm:\vspace{-0.5em}
\begin{equation}\label{0006}
c_i = -\sum_{j \in \mathcal{N}_i} (w_i - w_j - \Delta_{ij}), \quad \forall i \in \mathbb{Z}_1^n.
\end{equation}
Equation (\ref{0006}) can be written in a compact form as
\begin{equation}\label{0007}
\boldsymbol{c}(\boldsymbol{w}) = -L(\boldsymbol{w} - \boldsymbol{w}^*) = -L \tilde{\boldsymbol{w}},
\end{equation}
where $\boldsymbol{c}(\boldsymbol{w}) = (c_1(\boldsymbol{w}), \ldots, c_n(\boldsymbol{w}))^\top$, $L = L(\mathcal{G})$ is the Laplacian matrix of the graph $\mathcal{G}$, and $\tilde{\boldsymbol{w}} = \boldsymbol{w} - \boldsymbol{w}^*$. Then, the coordination vector field for the $i$-th robot, $\boldsymbol{\mathcal{X}_{\mathcal{P}_i}} : \mathbb{R}^{n+2} \to \mathbb{R}^3$ is designed as the weighted sum of the path-following vector field $^{\mathbf{pf}}\boldsymbol{\mathcal{X}_i}$ and the coordination component $^{\mathbf{co}}\boldsymbol{\mathcal{X}_i}$, given by:
\begin{equation}\label{0008}
\boldsymbol{\mathcal{X}_{\mathcal{P}_i}}(\bar{\boldsymbol{\xi}}_i, \boldsymbol{w}) = {}^{\mathbf{pf}}\boldsymbol{\mathcal{X}_i}(\bar{\boldsymbol{\xi}}_i) + k_c \, {}^{\mathbf{co}}\boldsymbol{\mathcal{X}_i}(\boldsymbol{w}),
\end{equation}
where $k_c > 0$ is a weighting parameter that adjusts the relative contribution of each component to ${}^{\mathbf{co}}\boldsymbol{\mathcal{X}_i}$. A larger $k_c$ results in faster coordination among the robots.

\subsection{Safe Collision Avoidance Vector Field}
\begin{figure}[t]
\centering
\subfigure[]{\includegraphics[width=0.24\linewidth]{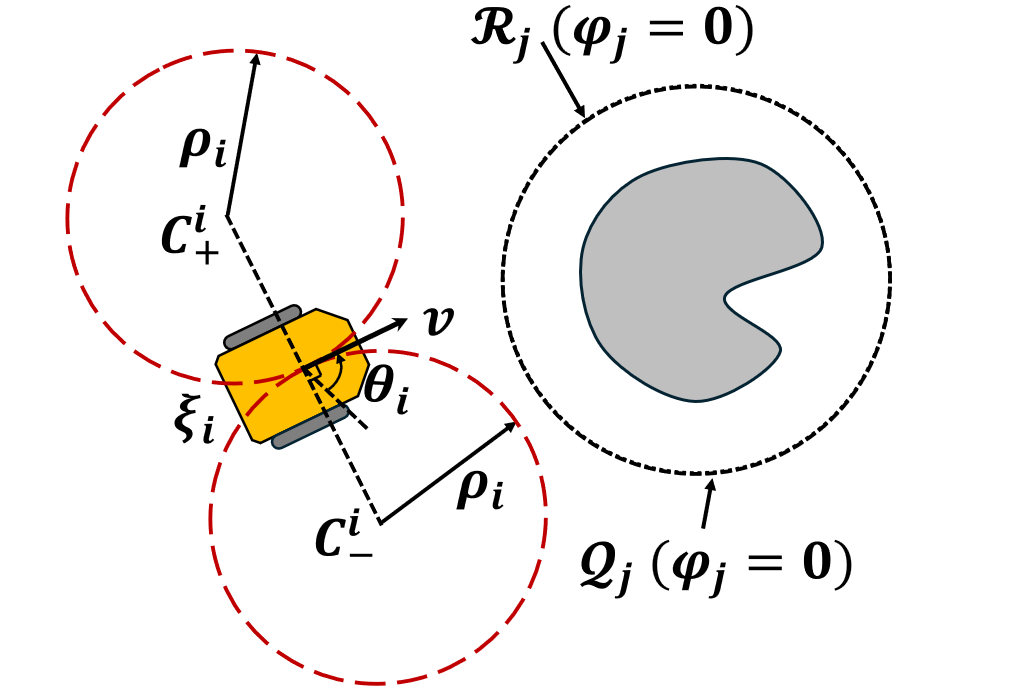}}
\subfigure[]{\includegraphics[width=0.24\linewidth]{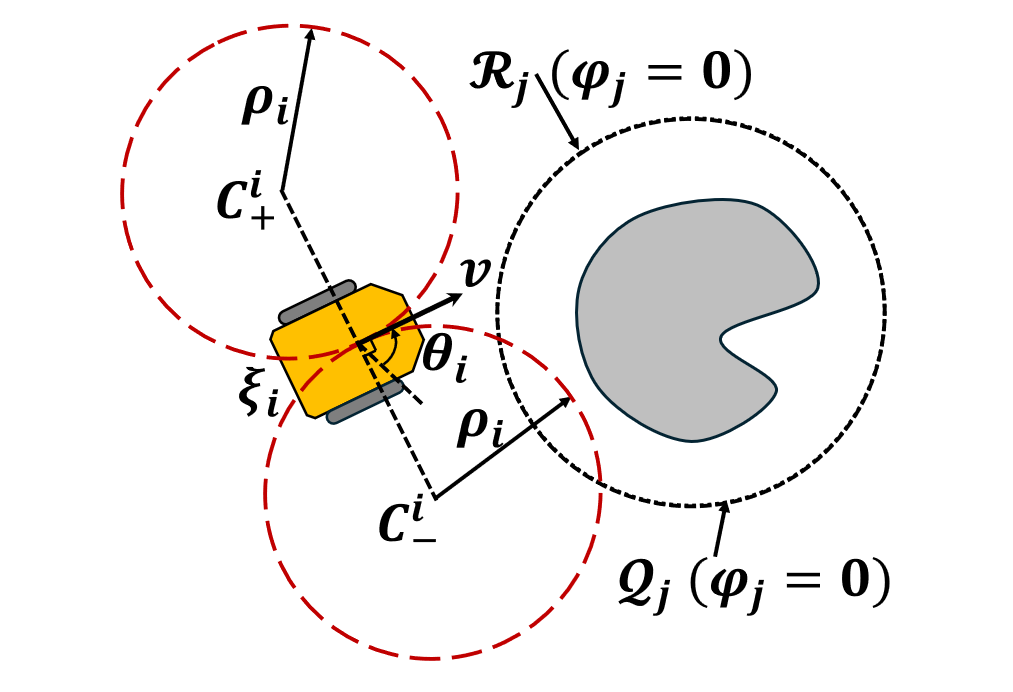}}
\subfigure[]{\includegraphics[width=0.24\linewidth]{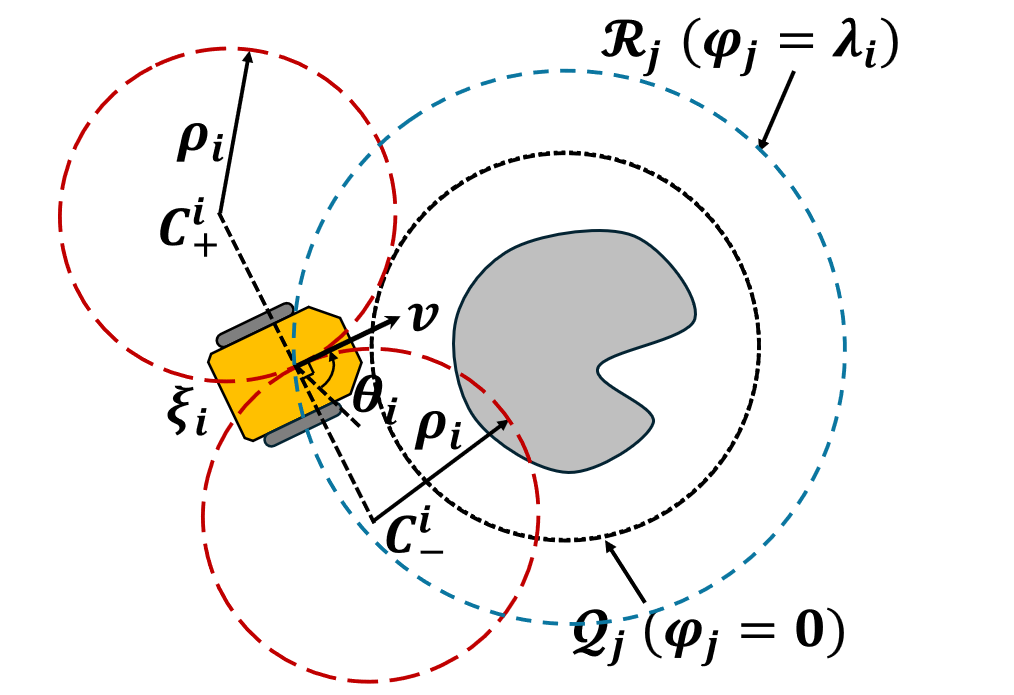}}
\subfigure[]{\includegraphics[width=0.24\linewidth]{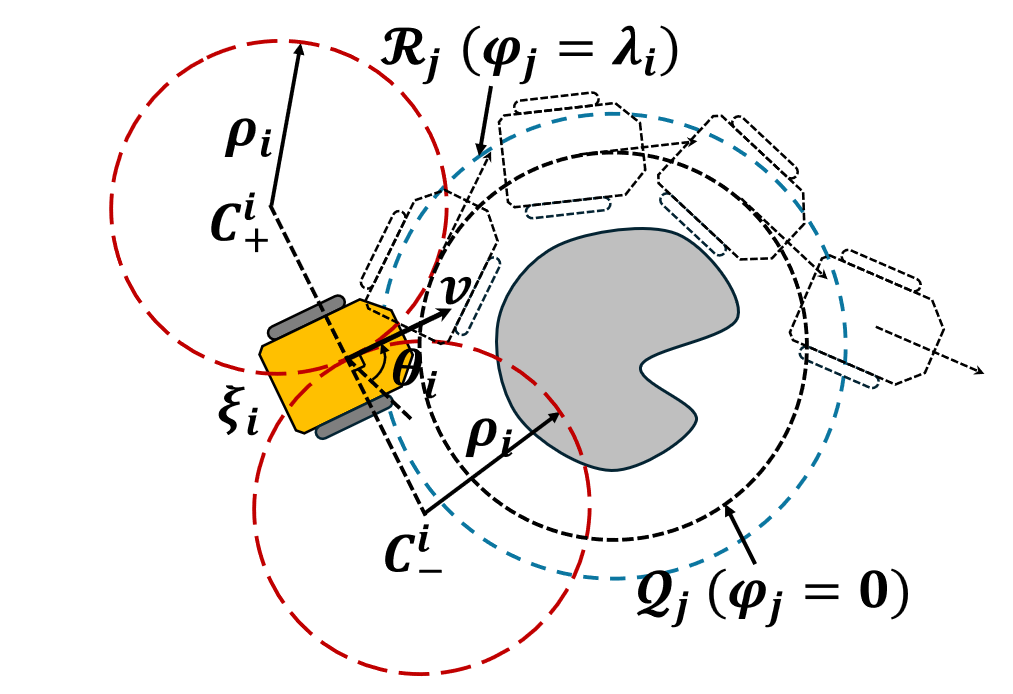}}
\caption{Illustration of obstacle avoidance with an adaptively adjustable reactive boundary. (a)~The limit circle is separated from the obstacle the obstacles, i.e., $\delta_+^{[i,j]}>0$, $\delta_-^{[i,j]}>0$, and $\gamma_j(\boldsymbol{\zeta}_i)=0$; (b)~The right limit circle intersects (or is tangent to) the obstacles while the left limit circle remains separated from the obstacle, i.e., $\delta_+^{[i,j]}>0$, $\delta_-^{[i,j]}\leq0$, and $\gamma_j(\boldsymbol{\zeta}_i)=0$; (c)~The right limit circle intersects the obstacles and the left limit circle is tangent to (or intersects) the obstacles, i.e., $\delta_+^{[i,j]}\leq0$, $\delta_-^{[i,j]}\leq0$, and $\gamma_j(\boldsymbol{\zeta}_i)=\varphi_j(\boldsymbol{\xi}_i)$, while $\delta_+^{[i,j]}>\delta_-^{[i,j]}$, indicating that the vector field guides the robot to avoid the obstacles by steering it laterally toward the left side relative to its motion direction; (d)~The ideal process of successful obstacle avoidance is shown, during which the reactive boundary continuously adapts dynamically.}
\label{fig404271}
\end{figure}

For the $i$-th robot with configuration \(\boldsymbol{\zeta}_i(t) = \left[\boldsymbol{\xi}_i, \theta_i \right]^\top\), and subject to a minimum turning radius constraint \(\rho_i\), a limit circle is defined as one of the circles that is tangent to the heading direction $\theta_i$, with a radius equal to \(\rho_i\), and passing the current position point $\boldsymbol{\xi}_i$. It corresponds to the robot's trajectory with the achievable maximum curvature $\bar{\kappa}_i=1/\rho_i$. The limit circles can be expressed as (see red circle in Fig.~\ref{fig404271}):
\begin{equation}\label{0009} 
C_\pm^i(t) = \left\{ \boldsymbol{p} \in \mathbb{R}^2 : \operatorname{dist}(\boldsymbol{p}, \boldsymbol{\xi}_i \pm \rho_i \mathbf{n}) = \rho_i \right\}, \quad \mathbf{n} = [-\sin\theta_i, \cos\theta_i]^\top
\end{equation}
where \(C_+^i(t)\) and \(C_-^i(t)\) denote the left and right limit circles of the $i$-th robot, $i \in \mathbb{Z}_1^n$, respectively. 

We define \(\delta_+^{[i,j]}(t)\) and \(\delta_-^{[i,j]}(t)\) to characterize the positional relations between the left(right) limit circles of the $i$-th robot and the obstacle \(\mathcal{O}_j, j \in \mathbb{Z}_1^m\)  at time $t$, as follows:\vspace{-0.5em}
\begin{align}
\delta_+^{[i,j]}(t) &= \operatorname{dist}(\boldsymbol{\xi}_{j}^o, C_+^i) - r_j, \label{4.11} \\
\delta_-^{[i,j]}(t) &= \operatorname{dist}(\boldsymbol{\xi}_{j}^o, C_-^i) - r_j. \label{4.12}
\end{align}
For notational convenience, the explicit dependence on time $t$ is henceforth omitted for all time-varying variables. When $\delta_\pm^{[i,j]} < 0$, the corresponding limit circle intersects the obstacle; when $\delta_\pm^{[i,j]} = 0$, it is tangent; and when $\delta_\pm^{[i,j]} > 0$, it is disjoint. Under fixed velocity constraints, if the robot starts with $\delta_\pm^{[i,j]} < 0$, 
collision is unavoidable regardless of the control input (see Fig.~\ref{fig404271}). 
Hence, the following assumption is imposed.

\begin{assumption}
The initial state satisfies $\delta_\pm^{[i,j]}(0) \geq 0$ for any $i \in \mathbb{Z}_1^n, j \in \mathbb{Z}_1^m$.
\end{assumption}

\begin{figure}[h]
\centering
\subfigure[]{\includegraphics[width=0.4\linewidth]{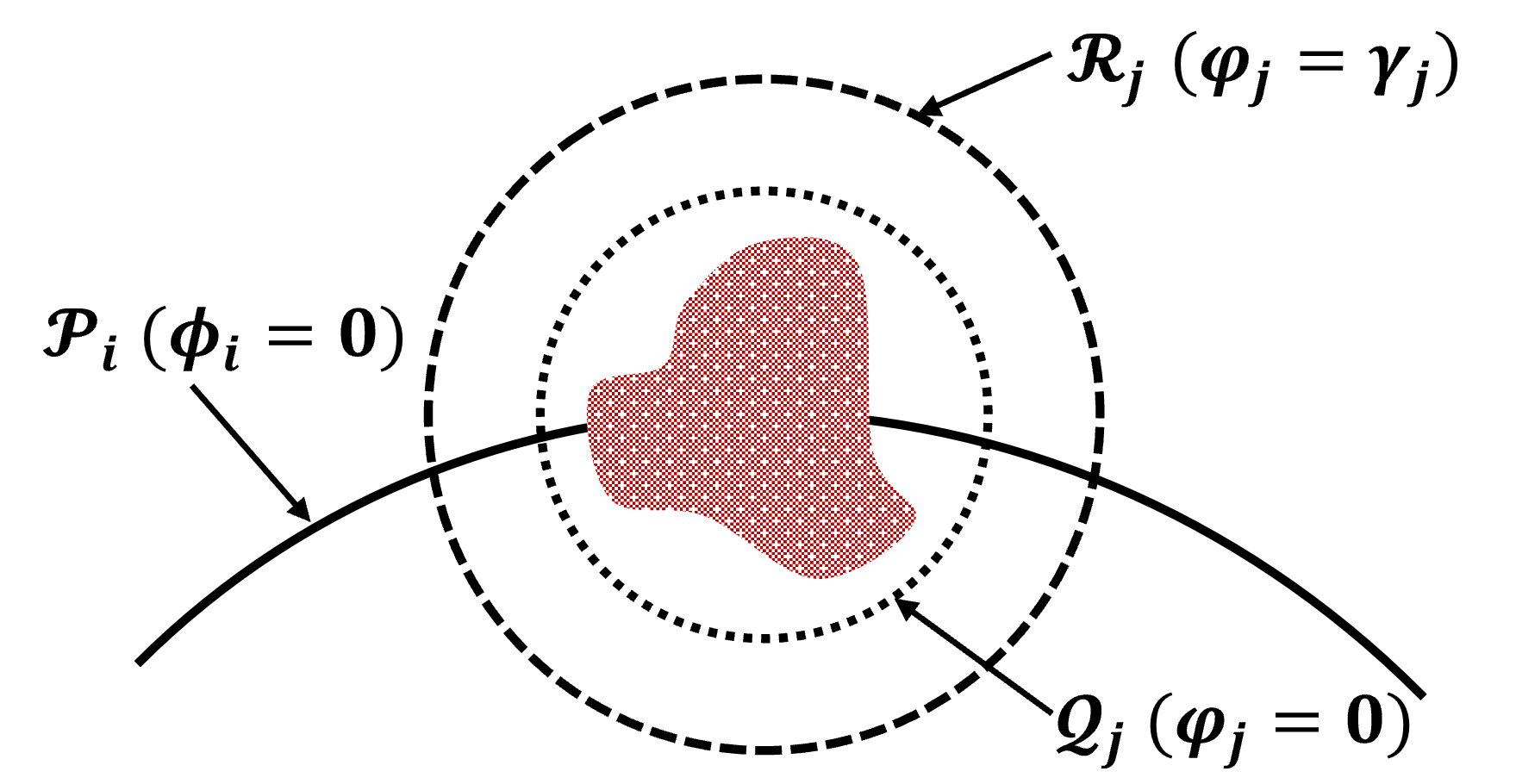}}
\subfigure[]{\includegraphics[width=0.25\linewidth]{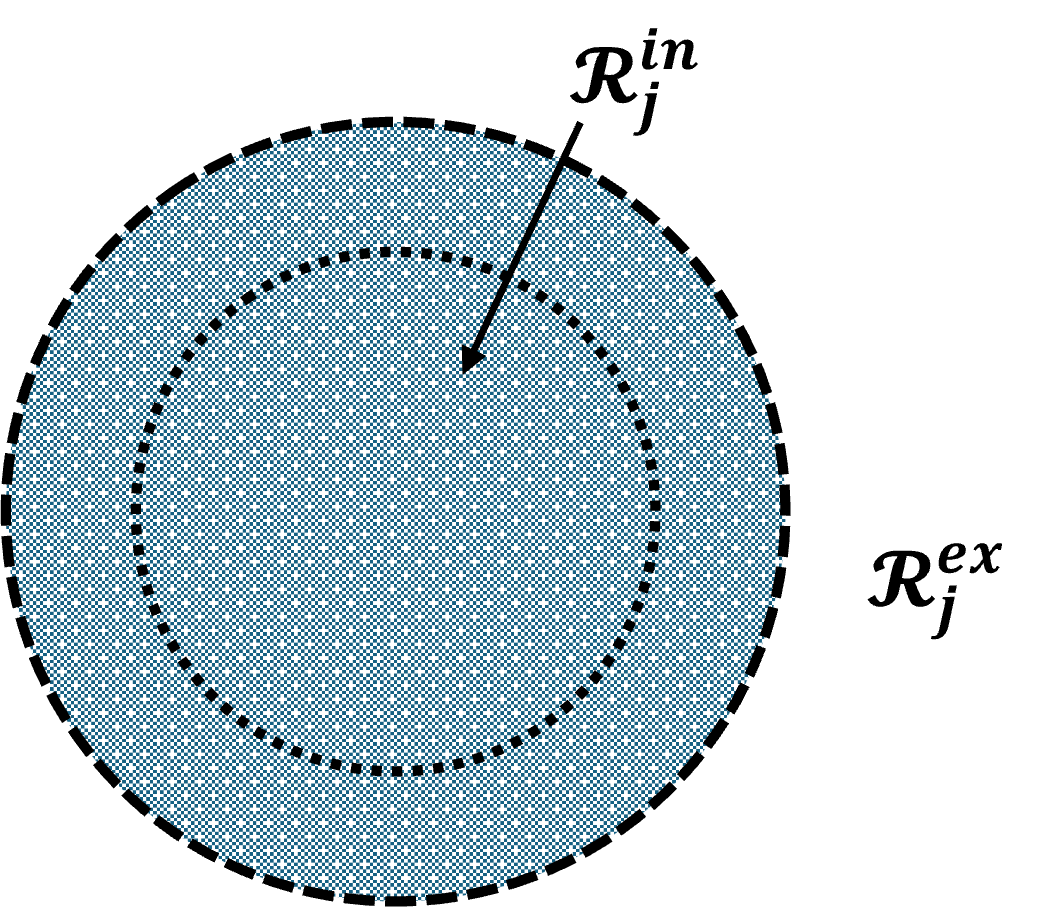}}
\subfigure[]{\includegraphics[width=0.25\linewidth]{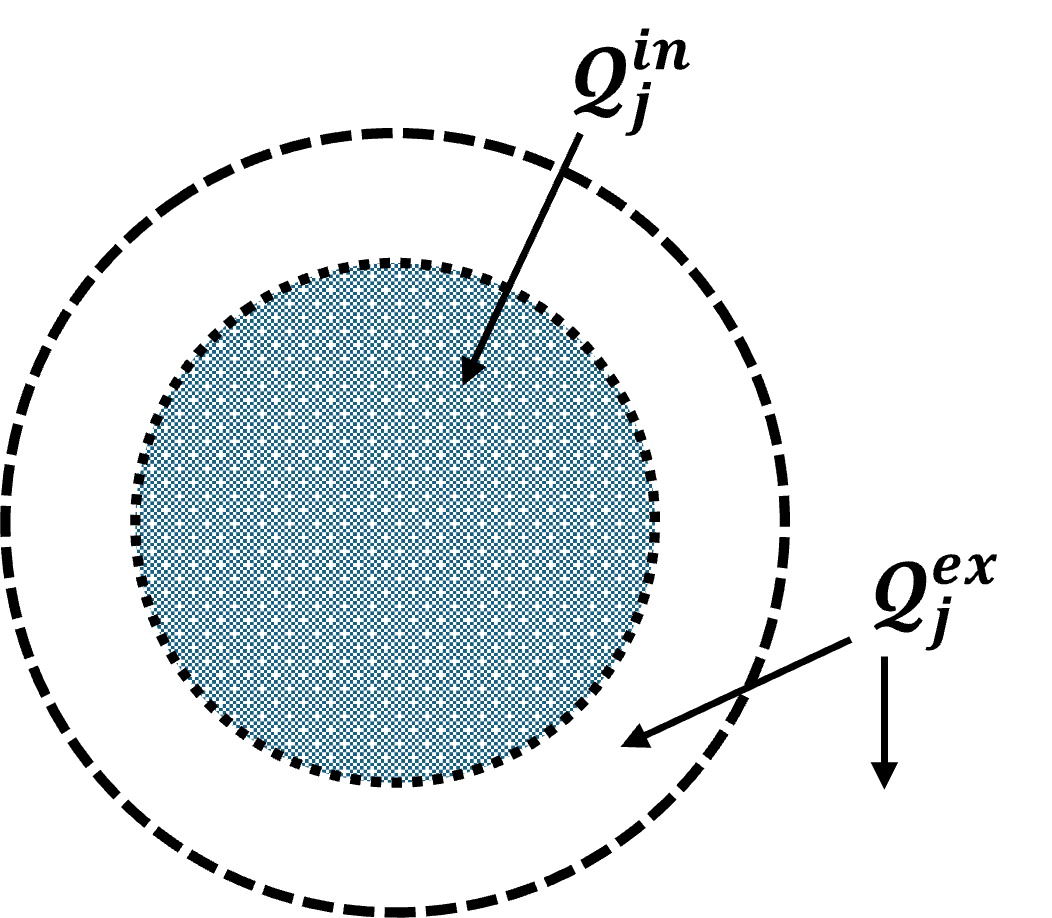}}
\caption{Illustration of region partitioning. (a)~Desired path \(\mathcal{P}_i\), reactive boundary \(\mathcal{R}_j\), repulsive boundary \(\mathcal{Q}_j\), and actual obstacles (red area); (b)~\(\mathcal{R}_j^{\mathrm{in}}\)(shaded area) and \(\mathcal{R}_j^{\mathrm{ex}}=\mathbb{R}^2 \setminus\mathcal{R}_j^{\mathrm{in}}\); (c)~\(\mathcal{Q}_j^{\mathrm{in}}\)(shaded area) and \(\mathcal{Q}_j^{\mathrm{ex}}=\mathbb{R}^2 \setminus \mathcal{Q}_j^{\mathrm{in}}\). The gray area represents the obstacle.}
\label{fig401}
\end{figure}

The desired path \(\mathcal{P}_i \subset \mathbb{R}^2\) is defined as the zero level set of a twice continuously differentiable function \(\phi_i: \mathbb{R}^2 \to \mathbb{R}\):\vspace{-0.5em}
\begin{equation}\label{0010}
\mathcal{P}_i = \{ \boldsymbol{p} \in \mathbb{R}^2 : \phi_i(\boldsymbol{p}) = 0 \}.
\end{equation}
We assume that $\phi_i$ is properly chosen such that  \(\mathcal{P}_i\) is a one-dimensional connected submanifold in \(\mathbb{R}^2\)\cite{yao:mar:lin:cao:21}. If \(\mathcal{P}_i\) is compact, it is homeomorphic to a circle; otherwise, it is homeomorphic to a real line interval \cite[Theorem~5.27]{Introduction}.

For the $i$-th robot and the $j$-th obstacle, where \(i \in \mathbb{Z}_1^n\) and \(j \in \mathbb{Z}_1^m\), we define a function \(\gamma_j: \mathbb{R}^3 \to \mathbb{R}\) that computes the level set of the obstacle’s reactive boundary based on the robot’s pose. It is defined as\vspace{-0.5em}
\begin{equation}\label{0012}
\gamma_j(t) =
\begin{cases}
0,  &\text{if}\; \sigma(\delta_+^{[i,j]}) \sigma(\delta_-^{[i,j]}) = 0, \\
\varphi_j(\boldsymbol{\xi}_i),  &\text{if} \; \sigma(\delta_+^{[i,j]}) \sigma(\delta_-^{[i,j]}) = 1,
\end{cases}
\end{equation}
where the characteristic function $\sigma:\mathbb{R} \to \mathbb{R}$ by $\sigma(x)=0$ for $x\in\left( 0, +\infty \right )$ and $\sigma(x)=1$ for $x\in\left( -\infty, 0 \right ]$. Furthermore, the repulsive boundary of the $j$-th obstacle is specified by a safety margin $s_{ij}>0$. Accordingly, two boundaries are defined surrounding the obstacle contour: the reactive boundary \(\mathcal{R}_j\) and the repulsive boundary \(\mathcal{Q}_j\) (see Fig.~\ref{fig401}), which are defined as
\begin{equation}\label{00008} 
\mathcal{R}_j = \{ \boldsymbol{p} \in \mathbb{R}^2 : \varphi_j(\boldsymbol{p}) = \gamma_j(t) \}, \quad 
\mathcal{Q}_j = \{ \boldsymbol{p} \in \mathbb{R}^2 : \varphi_j(\boldsymbol{p}) = s_{ij} \}.
\end{equation} 
where \(\mathcal{R}_j\) and \(\mathcal{Q}_j\) are one-dimensional compact connected submanifolds in \(\mathbb{R}^2\). By the Jordan Curve Theorem \cite{gowers2010princeton}, \(\mathcal{R}_j\) partitions the plane \(\mathbb{R}^2\) into a bounded open subset \(\mathcal{R}_j^{\mathrm{in}}\) (interior) and an unbounded open subset \(\mathcal{R}_j^{\mathrm{ex}}\) (exterior), satisfying \(\mathcal{R}_j = \partial \mathcal{R}_j^{\mathrm{in}} = \partial \mathcal{R}_j^{\mathrm{ex}}\), where \(\partial(\cdot)\) denotes the boundary operator. Similarly, \(\mathcal{Q}_j\) partitions the plane into \(\mathcal{Q}_j^{\mathrm{in}}\) and \(\mathcal{Q}_j^{\mathrm{ex}}\) with \(\mathcal{Q}_j = \partial \mathcal{Q}_j^{\mathrm{in}} = \partial \mathcal{Q}_j^{\mathrm{ex}}\), as illustrated in Fig.~\ref{fig401}. Note that the boundaries defined here are specific to the $i$-th robot.

For the adaptively adjustable reactive boundary \(\mathcal{R}_j\) defined in Eq. (\ref{00008}), the reactive vector field of the $i$-th robot, 
\(\boldsymbol{\mathcal{X}_{\mathcal{R}_j}} : \mathbb{R}^3 \to \mathbb{R}^2\) is defined as:\vspace{-0.5em}
\begin{align}
\boldsymbol{\mathcal{X}_{\mathcal{R}_j}}(\boldsymbol{\zeta}) &= \left( 2\sigma(\delta_+^{[i,j]} - \delta_-^{[i,j]}) - 1 \right) E \nabla \varphi_j(\boldsymbol{\xi}) - k_{r_j} \varphi_j(\boldsymbol{\xi}) \nabla \varphi_j(\boldsymbol{\xi}), \label{0013}
\end{align}
where the matrix \(E = \begin{bmatrix} 0 & -1 \\ 1 & 0 \end{bmatrix}\) represents a \(90^\circ\) rotation, \(k_{r_j} > 0\) is a positive gain, and \(\nabla(\cdot)\) denotes the gradient with respect to \(\boldsymbol{\xi}\). It is noteworthy that the term $2\sigma(\delta_+^{[i,j]} - \delta_-^{[i,j]})$ provides a tunable mechanism to regulate the direction of the $j$-th obstacle avoidance for the $i$-th robot.

\begin{remark}
Under Assumption~1, suppose that at a certain time instant $t^*$ the condition $\sigma(\delta_+^{[i,j]}) \sigma(\delta_-^{[i,j]}) = 1$ holds. Without loss of generality, assume that $\delta_+^{[i,j]} \geq \delta_-^{[i,j]}$. At this moment, starting from $t^*$, the direction of vector field (\ref{0013}) is given by $E\frac{\boldsymbol{\xi}_j^o-\boldsymbol{\xi}_i}{\|\boldsymbol{\xi}_j^o-\boldsymbol{\xi}_i\|}$, which forces the robot to turn left. If the controller is designed such that the robot can maneuver with the minimum turning radius, it can precisely avoid the obstacle. 

\end{remark}

For inter-robot collision avoidance,  robots are treated as moving obstacles to implement collision avoidance among them. Namely, for robots indexed by \(i_1\) and \(i_2\) with \(1 \leq i_1 < i_2 \leq n\), robot \(i_2\) regards robot \(i_1\) as an obstacle and defines a corresponding reactive boundary \(\mathcal{R}_{i_1}\) and a repulsive boundary \(\mathcal{Q}_{i_1}\) for collision avoidance control.

Furthermore, for any reactive boundary \(\mathcal{R}_j\) and repulsive boundary \(\mathcal{Q}_j, j \in \mathbb{Z}_1^{m+n}\), we design smooth zero-in and zero-out bump functions \(\sqcup_{\mathcal{Q}_j}, \sqcap_{\mathcal{R}_j} : \mathbb{R}^2 \to [0, \infty)\) as follows:\vspace{-0.5em}
\begin{equation}\label{0014}
\sqcup_{\mathcal{Q}_j}(\xi) =
\begin{cases}
0, & \xi \in \overline{\mathcal{Q}_j^{\mathrm{in}}}, \\
a_j(\xi), & \xi \in \mathcal{Q}_j^{\mathrm{ex}},
\end{cases}
\quad
\sqcap_{\mathcal{R}_j}(\xi) =
\begin{cases}
0, & \xi \in \overline{\mathcal{R}_j^{\mathrm{ex}}}, \\
b_j(\xi), & \xi \in \mathcal{R}_j^{\mathrm{in}},
\end{cases}
\end{equation}
where \(a_j: \mathcal{Q}_j^{\mathrm{ex}} \to (0,\infty)\) and \(b_j: \mathcal{R}_j^{\mathrm{in}} \to (0,\infty)\) are bounded. These functions are used to smoothly "blend" vector fields in different areas as shown later. 

After introducing the aforementioned vector fields and bump functions, the final safety cooperative vector field for the \(i\)-th robot, \(\boldsymbol{\mathfrak{X}_i} : \mathbb{R}^{n+3} \to \mathbb{R}^3\), is defined by combining equations (\ref{0008}), (\ref{0013}), and (\ref{0014}) as follows:
\begin{equation}\label{0015}
\boldsymbol{\mathfrak{X}_i}(\bar{\boldsymbol{\xi}}_i, \boldsymbol{\zeta}_i, \boldsymbol{w}) = 
\prod_{j \in \mathbb{Z}_1^{m+n}} \sqcup_{\mathcal{Q}_j}(\boldsymbol{\xi}_i) \, \hat{\boldsymbol{\mathcal{X}}}_{\mathcal{P}_i}(\bar{\boldsymbol{\xi}}_i) 
+ \sum_{j \in \mathbb{Z}_1^{m+n}} \sqcap_{\mathcal{R}_j}(\boldsymbol{\xi}_i) \, \hat{\boldsymbol{\mathcal{X}}}_{\mathcal{R}_j}(\boldsymbol{\zeta}_i)
\end{equation}

\begin{remark}
   Although Equation (\ref{0015}) shares a structural resemblance to the vector field formulation in \cite{yao:lin:and:cao:22}, its functional role is fundamentally different. In Equation (\ref{0015}), the path-tracking mechanism is reformulated by substituting the original guiding vector field with a cooperative vector field, thereby enabling coordinated motion. Moreover, by introducing adaptively adjustable reactive boundary, the collision-avoidance vector field is systematically extended to accommodate multi-robot scenarios while preserving physical feasibility. Last but not least, the vector field in \cite{yao:lin:and:cao:22} does not consider the robot trajectory curvature constraints, while our proposed vector field (\ref{0015}) can deal with these constraints. 
\end{remark}

\subsection{Control Law Design}
The detailed control design is as follows. First, we define the saturation function $\mathrm{Sat}_a^b:\mathbb{R}\to\mathbb{R}$ by $\mathrm{Sat}_a^b(x)=x$ for $x\in[a,b]$, $\mathrm{Sat}_{a}^{b}(x)=a$ for $x\in(-\infty,a)$ and $\mathrm{Sat}_{a}^{b}(x)=b$ for $x\in(b,\infty)$, where $a,b\in\mathbb{R}$, $a<b$ are some constants. The saturation function Sat$_a^b$ is Lipschitz continuous. To ensure that the heading direction of the robot is aligned with that of the vector field, we define \vspace{-0.5em}
\begin{equation}
    \theta_i^d = \arctan\frac{\boldsymbol{\mathfrak{X}}_{i2}}{\boldsymbol{\mathfrak{X}}_{i1}},
\end{equation}
where $\boldsymbol{\mathfrak{X}}_{i1}$ and $\boldsymbol{\mathfrak{X}}_{i2}$ denote the components of the desired vector field at the $i$-th robot's position. For any nonsingular point of the vector field, the angular velocity control input of the $i$-th robot is given by  \vspace{-0.5em}
\begin{equation} \label{kzl}
    \dot{\theta}_i = \omega_i = \mathrm{Sat}_a^b \left( k_\theta \left( \theta_i^d - \theta_i \right) \right),
\end{equation}
where $k_{\theta} > 0$ is a constant gain, and $a = -\bar{u}_i$, $b = \bar{u}_i$.

\begin{remark}
If $a_j \to 0,b_j \to 1$, and $k_{\theta}$ is sufficiently large, the robot can achieve obstacle avoidance and path following under vector field (\ref{0015}) and controller (\ref{kzl}). The control law (\ref{kzl}) is rather coarse, in future work, we will present a more detailed control strategy (similar to \cite[eq.~(24a)]{yao:mar:lin:cao:21}).
\end{remark}
    
\section{Simulation and Experiment}\label{si}
We adopt a ring graph as the communication topology, such that each robot is only allowed to communicate with its two immediate neighbors. This simple communication graph significantly reduces both communication and computational burden.

\subsection{Simulation}
\begin{figure}[h]
\centering
\subfigure[]{\label{fig4.1:subfig:a}
\includegraphics[width=0.31\linewidth]{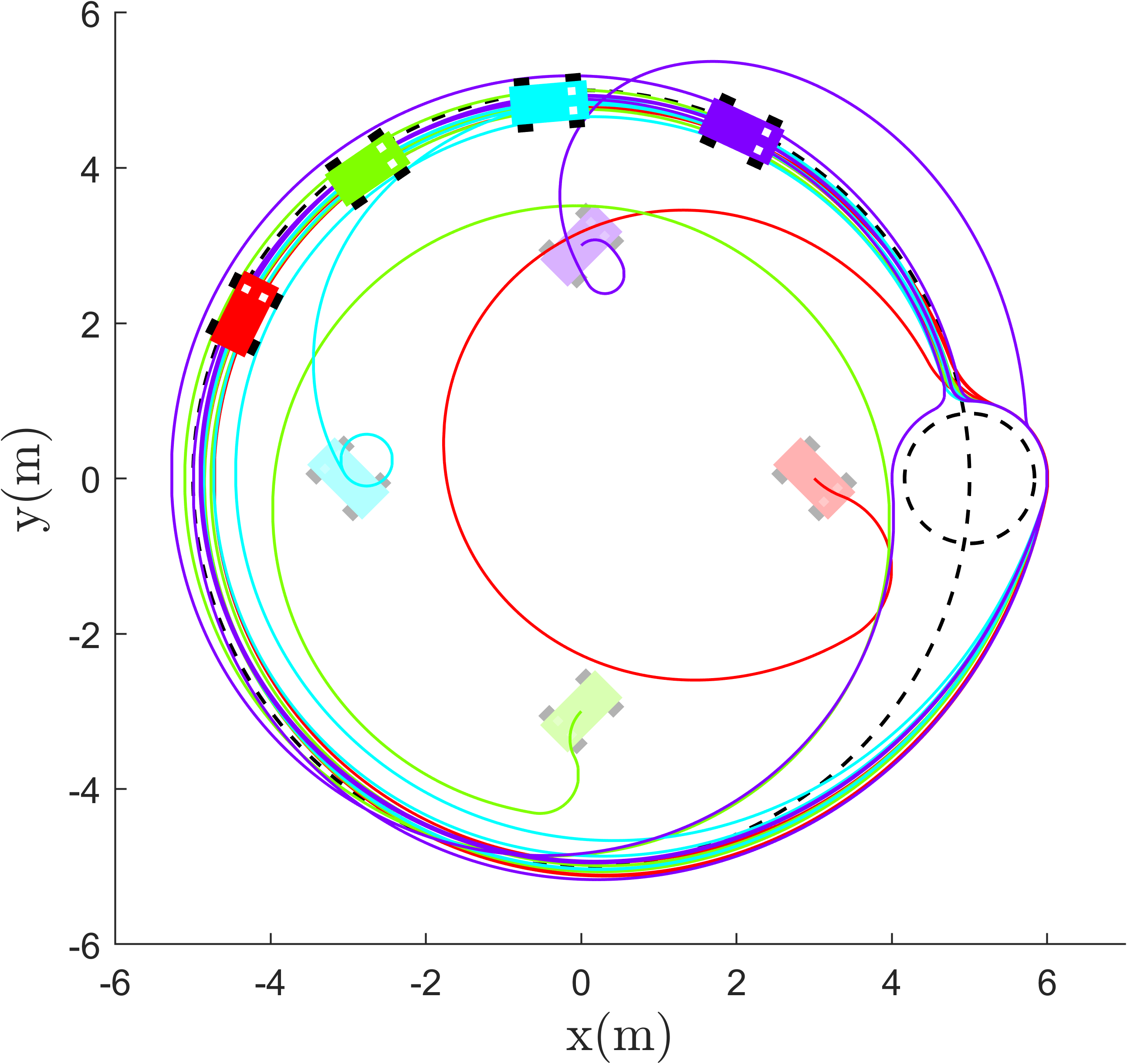}} 
\subfigure[]{\label{fig4.1:subfig:b}
\includegraphics[width=0.31\linewidth]{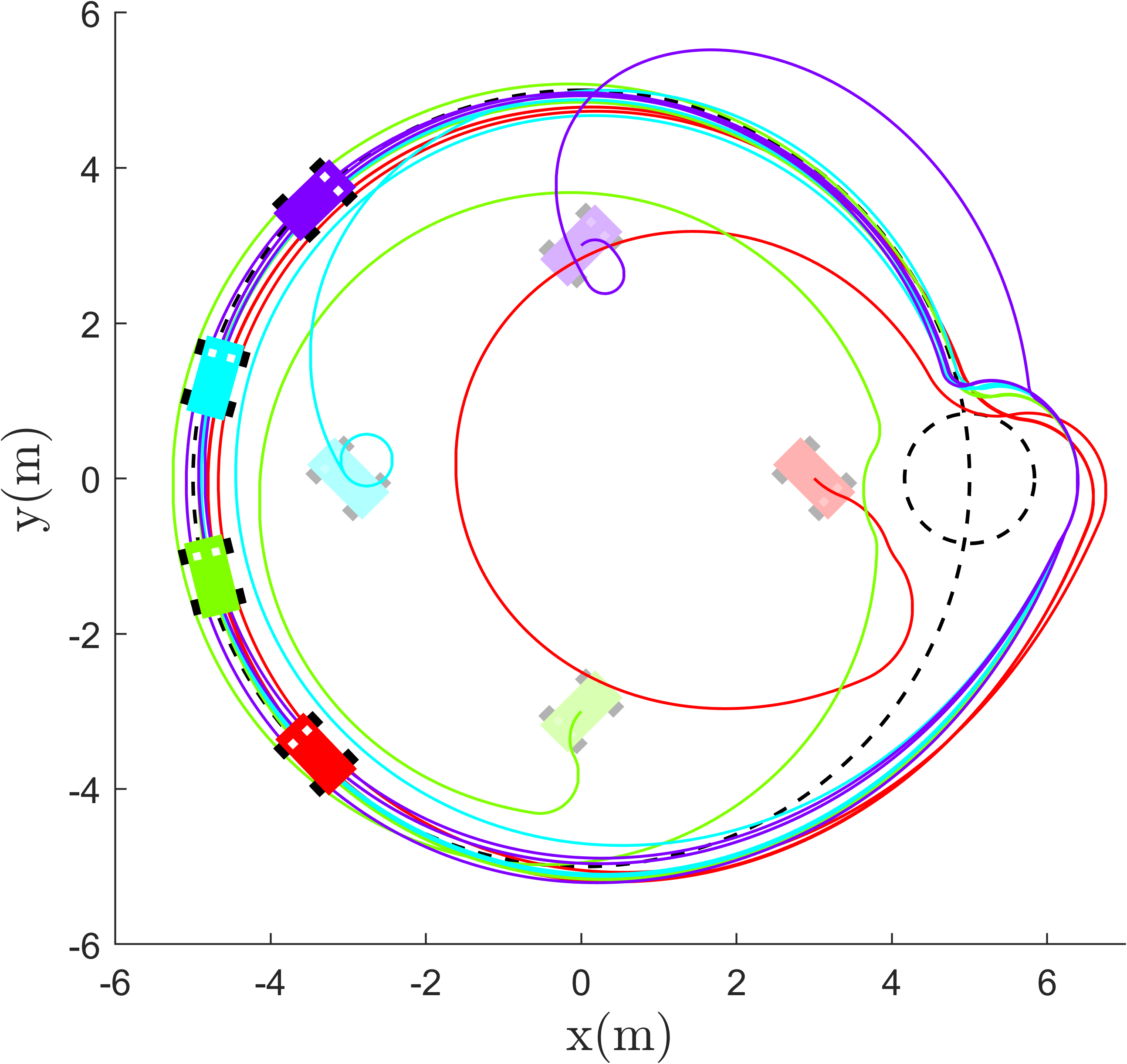}} 
\subfigure[]{\label{fig4.1:subfig:c}
\includegraphics[width=0.31\linewidth]{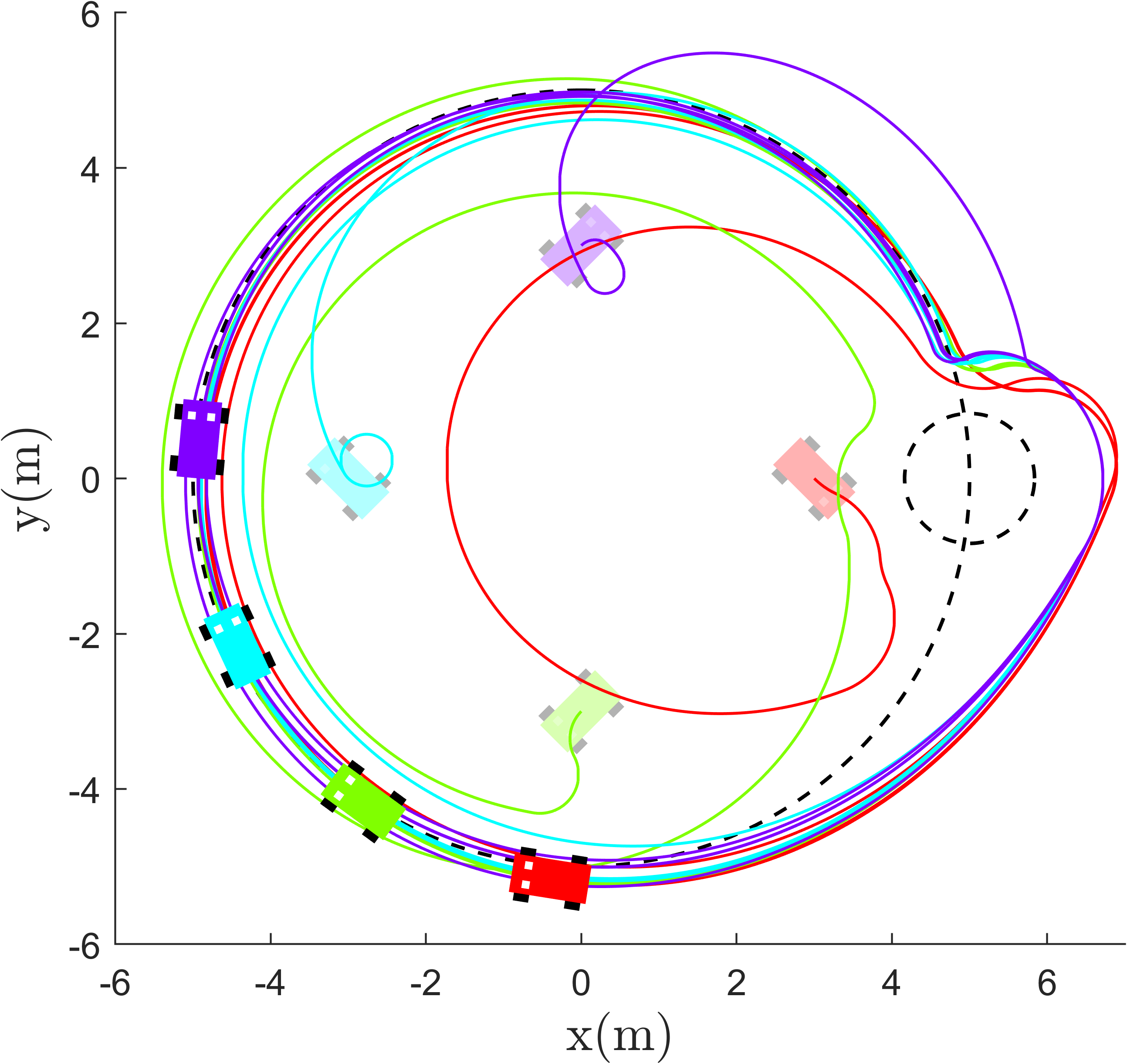}}

\subfigure[]{\label{fig4.1:subfig:d}
\includegraphics[width=0.31\linewidth]{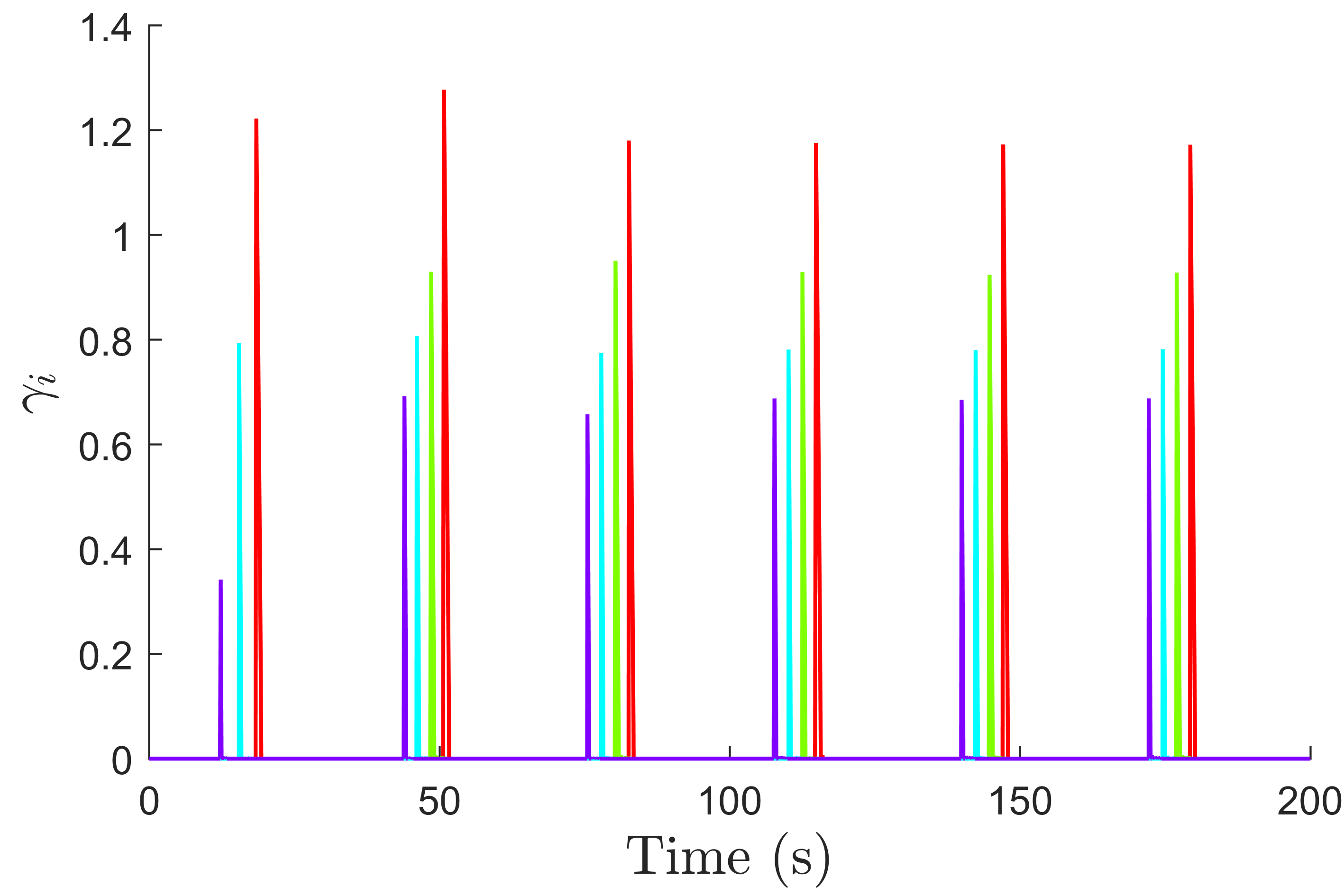}}
\subfigure[]{\label{fig4.1:subfig:e}
\includegraphics[width=0.31\linewidth]{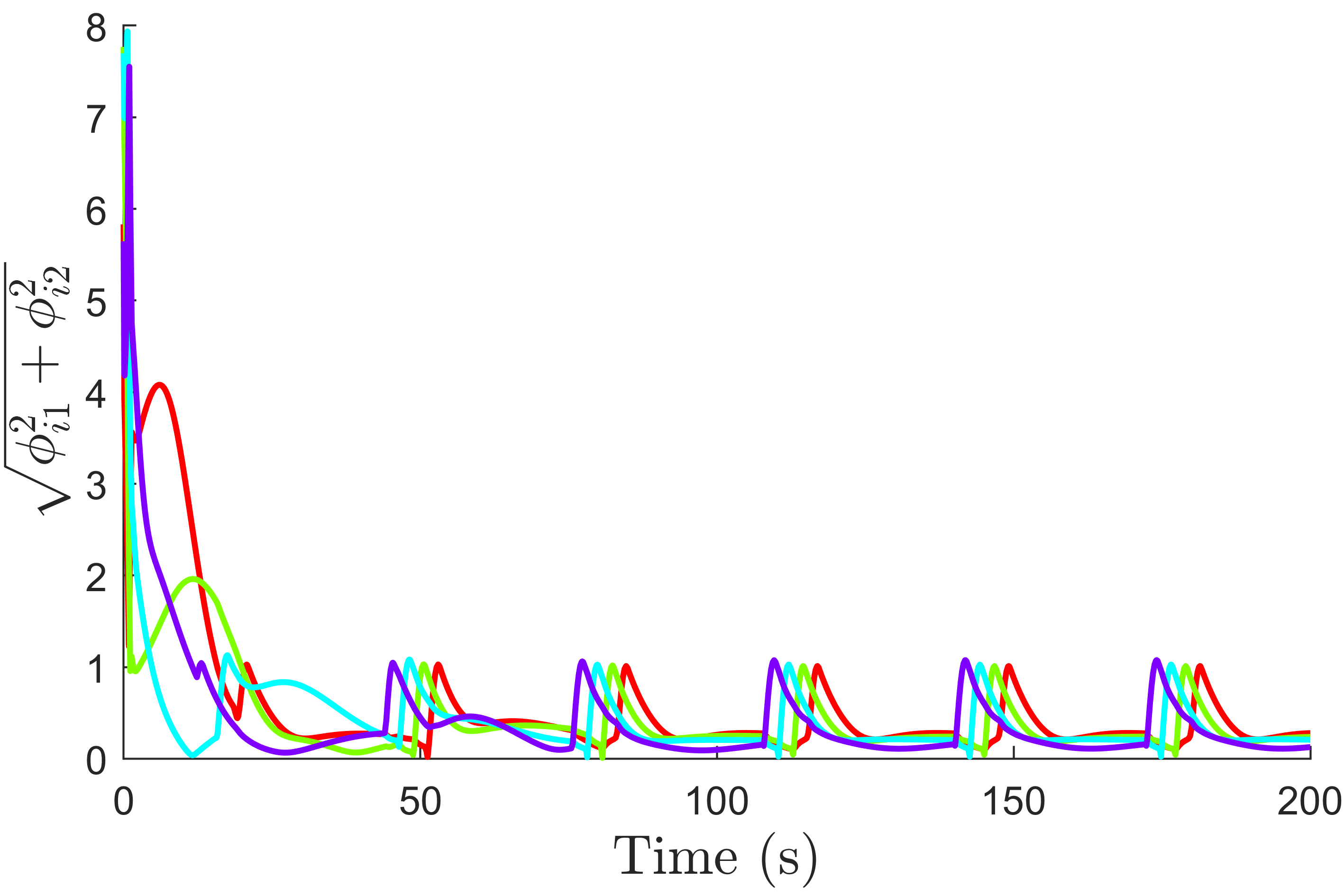}}
\subfigure[]{\label{fig4.1:subfig:f}
\includegraphics[width=0.31\linewidth]{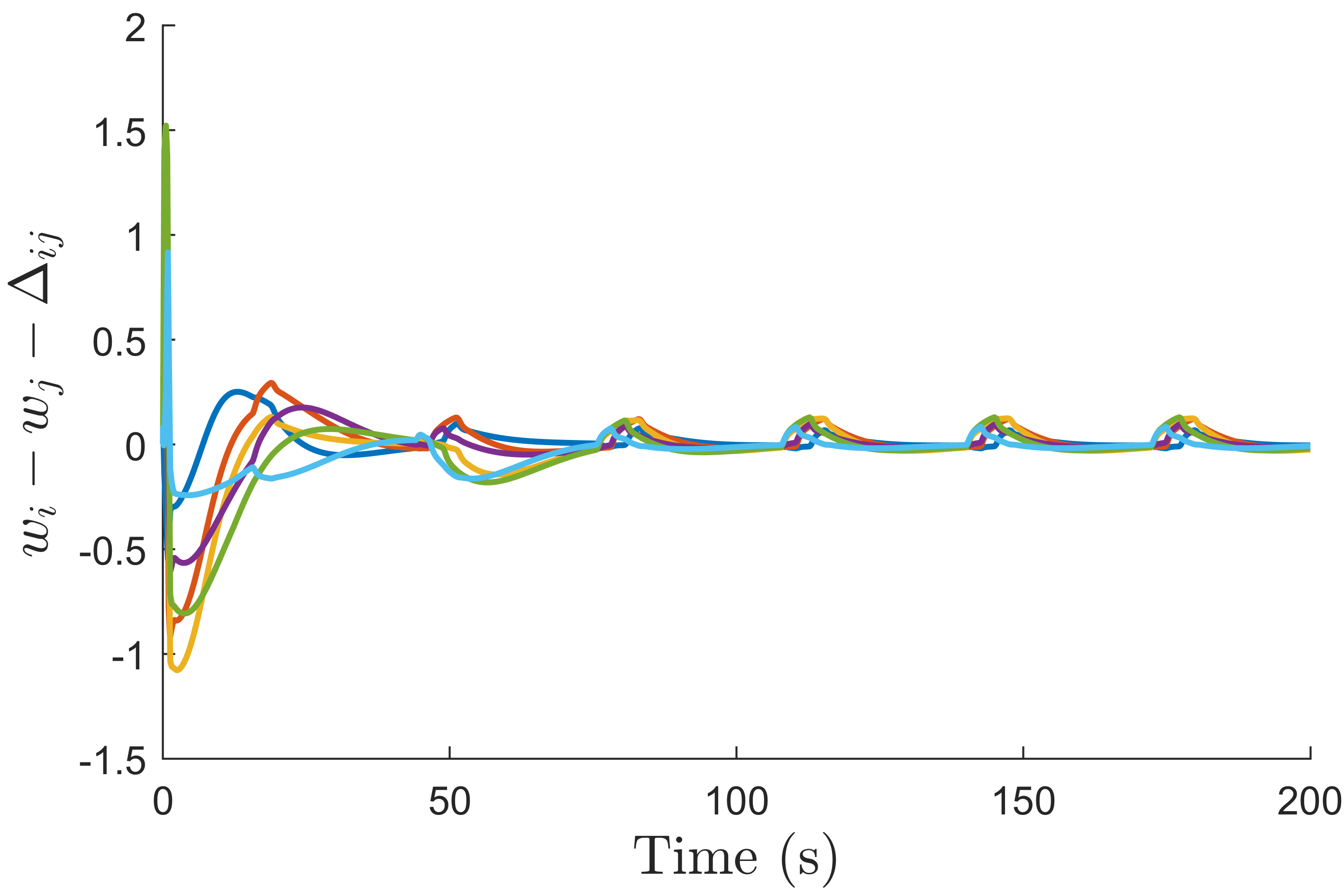}}

\caption{
Simulation results with the desired path parameterized as \(x_{i1} = 5\sin(\omega_i)\) and \(x_{i2} = 5\cos(\omega_i)\), using gains \(k_{i1} = k_{i2} = 1\) and coordination gain \(k_c = 10\), for \(i \in \mathbb{Z}_1^n\). The desired parameter offsets \(\Delta_{ij}\) are defined based on the reference trajectories \( \omega_i^*(t) = 0.5\, i,\; i \in \mathbb{Z}_1^n\). Additional parameters are set as \(k_{r_i} = 1\), \(k_{\theta} = 100\), \(a_i(\boldsymbol{\xi}) = 0.1\), \(b_i(\boldsymbol{\xi}) = 0.9\) for \(i \in \mathbb{Z}_1^n\). The implicit function for the repulsive boundary is given by \(\varphi_1 = (x - 5)^2 + y^2 - 1\). (a)~Adaptive reactive boundary; (b)~\(\gamma_i \equiv 1\); (c)~\(\gamma_i \equiv 2\): robot trajectories are shown, black dashed curves represent repulsive boundaries and desire path; (d)~Temporal evolution of the adaptive reactive boundary \(\gamma_i, i \in \mathbb{Z}_1^n\); (e)~Path tracking errors \(\sqrt{\phi_{i1}^2 + \phi_{i2}^2},\; i \in \mathbb{Z}_1^n\); (f)~Coordination errors \(\omega_i - \omega_j - \Delta_{ij}\) for \(i,j \in \mathbb{Z}_1^n,\; i < j\).
}
\label{fig4455}
\end{figure}

In the simulation experiment, we let \(n=4\) heterogeneou robots move along a circular trajectory. As shown in Fig.~\ref{fig4455}, all robots successfully follow the circular path while maintaining the desired relative positions and successfully avoid obstacles on the path. The speeds of the four robots are set to \(v=1\), with maximum angular velocities of 1, 2, 3, and 4 respectively.

\begin{remark}
    Since obstacles affect coordination among robots and the path-following performance, the adaptive reaction boundary $\gamma_i$, the path-tracking error, and the motion coordination error all exhibit approximately periodic fluctuations. Under the influence of the adaptive reactive boundary, the robots neither collide with obstacles due to an excessively small $\gamma_i$ nor generate redundant detours due to an excessively large $\gamma_i$, but instead achieve better obstacle avoidance behavior.
\end{remark}

\begin{figure}[h]
\centering

\subfigure[$t=0s$]{\label{qqqq}
\includegraphics[width=0.33\linewidth]{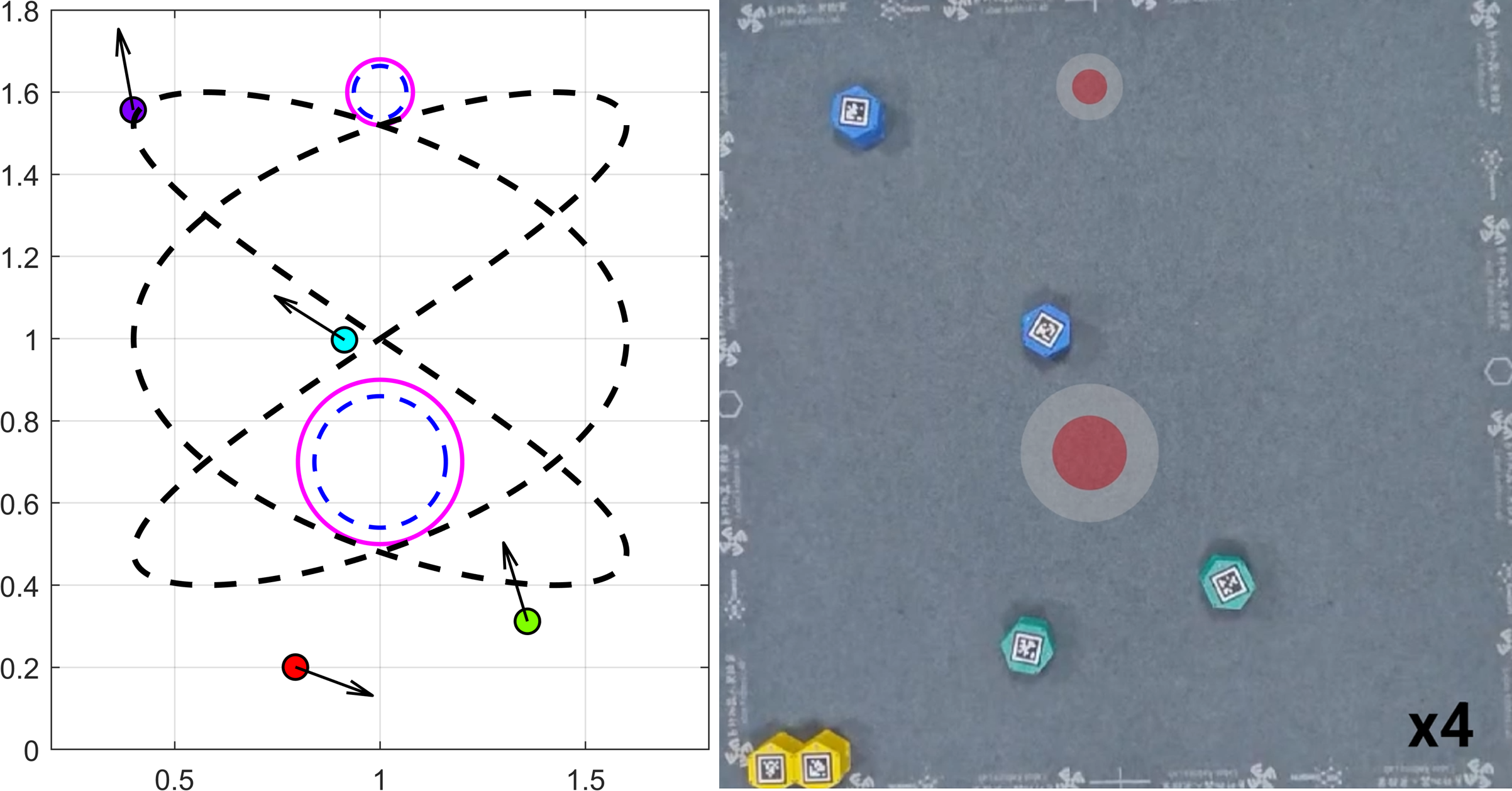}}
\subfigure[$t=54.75s$]{\label{wwww}
\includegraphics[width=0.33\linewidth]{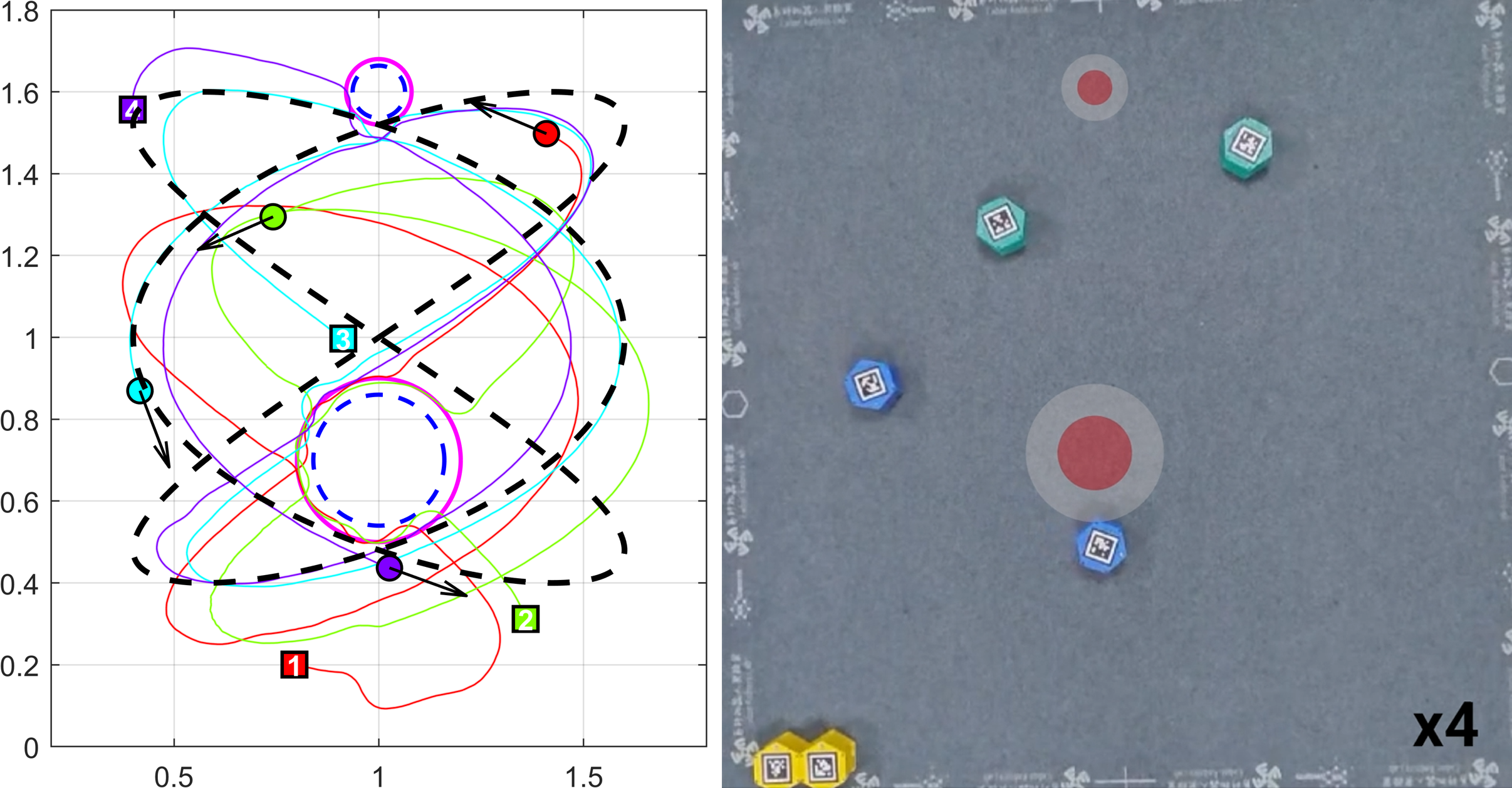}}
\subfigure[]{\label{figexp:subfig:c}
\includegraphics[width=0.27\linewidth]{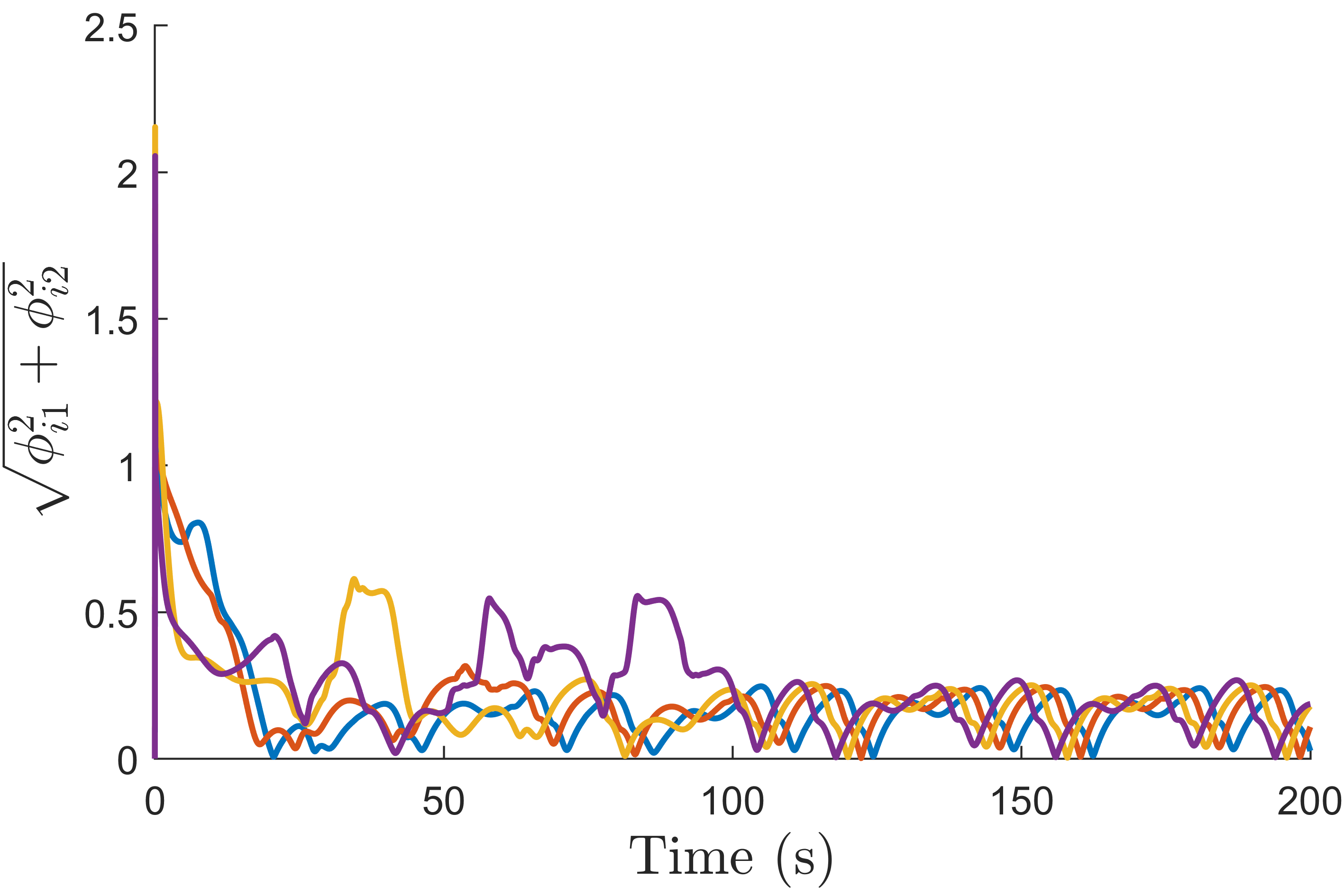}}

\subfigure[$t=129.14s$]{\label{sssss}
\includegraphics[width=0.33\linewidth]{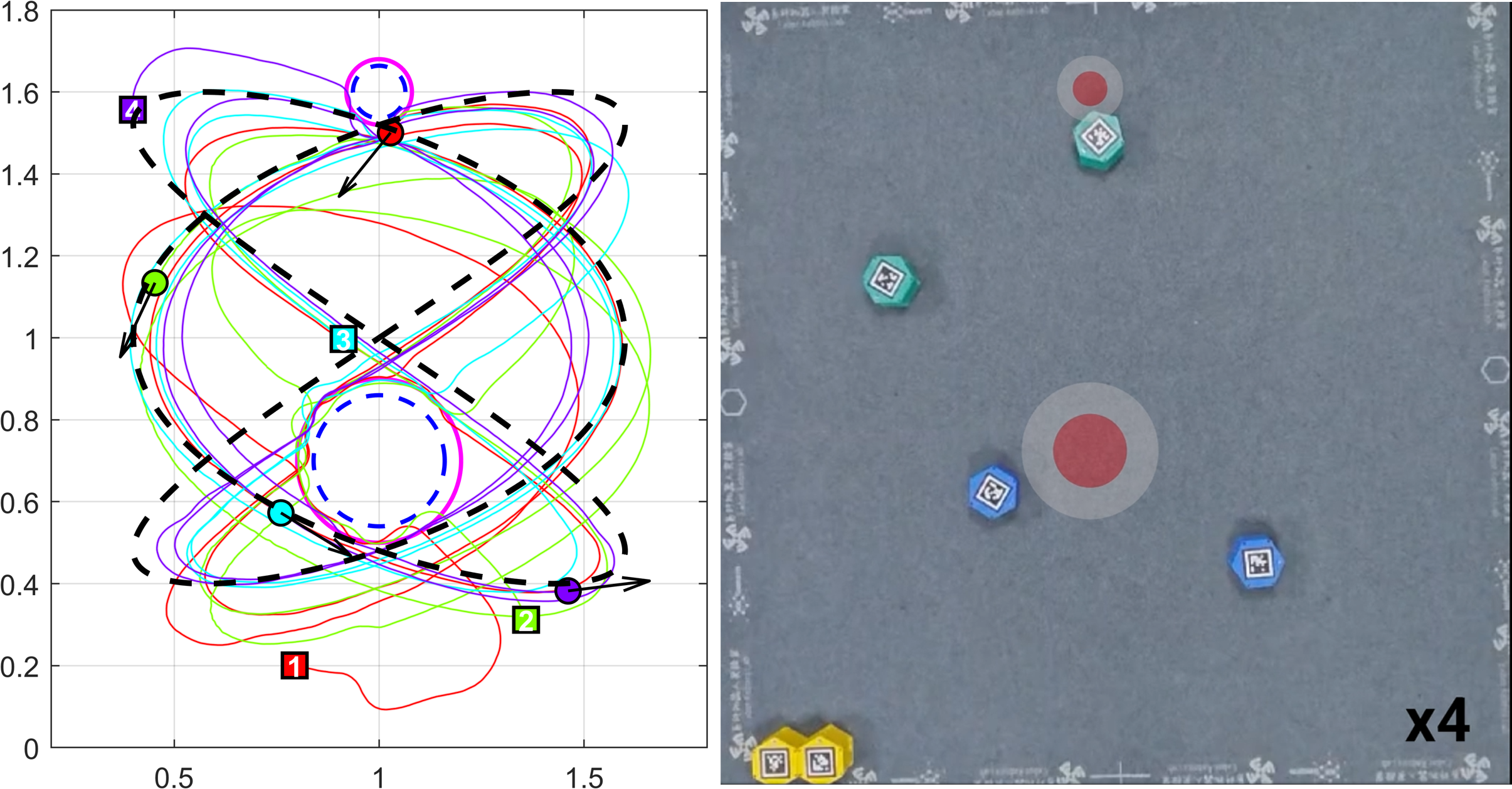}}
\subfigure[$t=177.95s$]{\label{ddddd}
\includegraphics[width=0.33\linewidth]{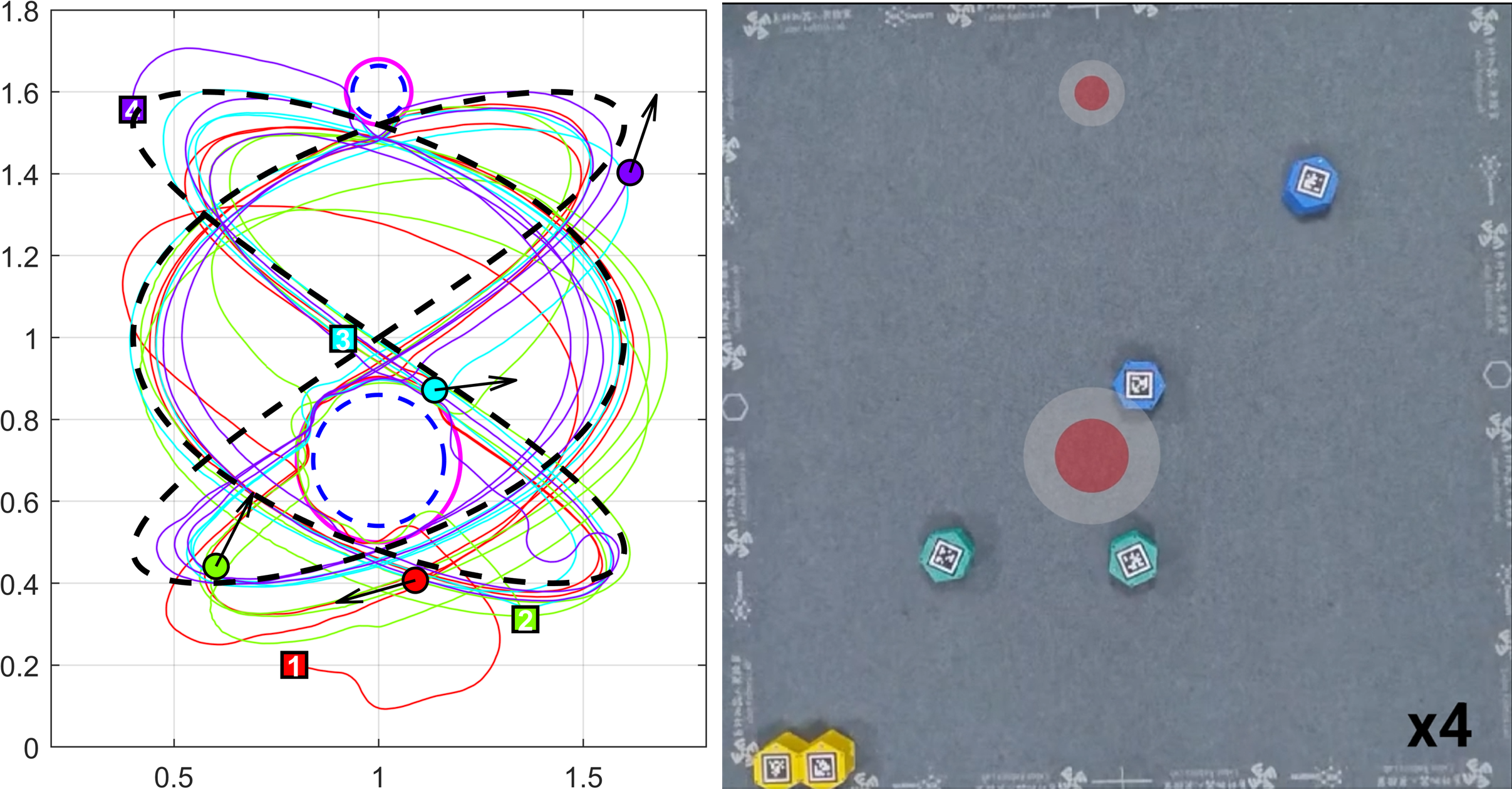}}
\subfigure[]{\label{figexp:subfig:f}
\includegraphics[width=0.27\linewidth]{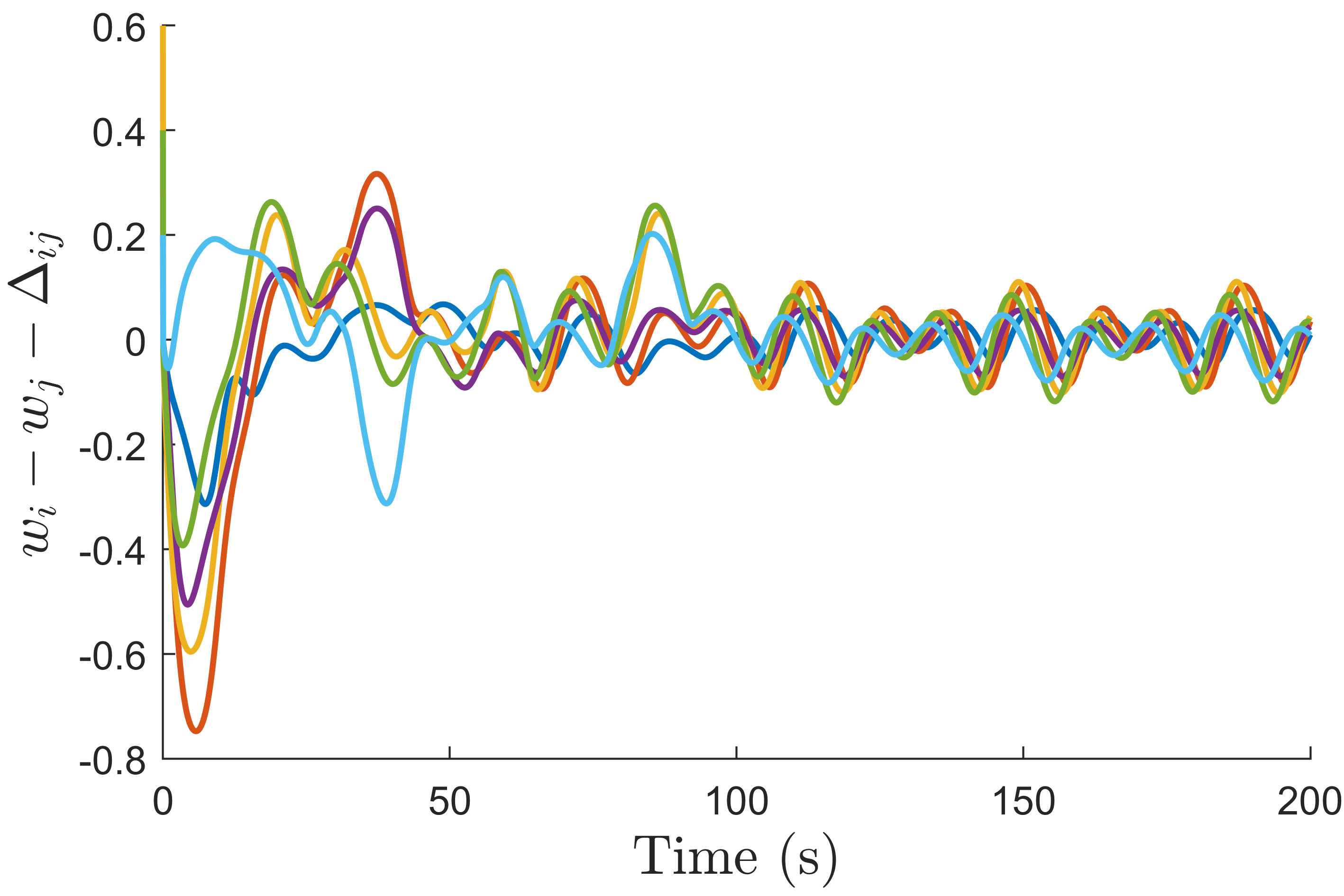}}

\caption{(a)(b)(d)(e) Snapshots from the physical experiment. The left subfigures illustrate the experimental trajectories at the corresponding time instants, while the right subfigures show simultaneous images of the real system captured by cameras. Small circles denote the current positions of the robots, small squares denote their initial positions, black arrows indicate the current heading directions, curves with different colors represent the trajectories of the robots, black dashed circles indicate the actual obstacles, and pink solid circles indicate the repulsive boundaries. (c)(f) Experimental data from the physical platform. The left figure shows the path-following errors \(\sqrt{\phi_{i1}^2 + \phi_{i2}^2}\), \(i \in \mathbb{Z}_1^n\); the right figure shows the coordination errors \(\omega_i - \omega_j - \Delta_{ij}\) for \(i, j \in \mathbb{Z}_1^n, i < j\).}
\label{qwer}
\end{figure}

\subsection{Multi-Robot Platform Experiment}
We deployed our vector field algorithm on a multi-robot experimental platform to test the formation maintenance and safe obstacle avoidance performance of multiple robots along a closed trajectory. The experimental setup is as follows: The desired path is a Lissajous curve parameterized by $x_{i1} = 0.6 \sin(n_x w_{i1}) + 1$ and $x_{i2} = 0.6 \sin(n_y w_{i1}) + 1$. Here, the coefficients are set as \(n_x = 3\) and \(n_y = 2\). Since the ratio \(n_x / n_y\) is rational, the curve forms a closed path. The parameters of the cooperative guidance vector field are set as \(k_{i1} = k_{i2} = 1, k_c = 1\), for \(i \in \mathbb{Z}_1^n\). The desired parameter differences \(\Delta_{ij}\) are constructed based on the reference trajectory \(w_i^*(t) = 0.5 i\) for $i \in \mathbb{Z}_1^n$. The parameters for the composite vector field are given by \(k_p = 1\), \(k_r = 1\), \(k_{\theta} = 5\), \(a_i(\boldsymbol{\xi}) = 0.1\), \(b_i(\boldsymbol{\xi}) = 0.9\), and \(k_s = 1.25\), which represents the ratio of the repulsive boundary radius to the actual obstacle radius. The implicit functions for the repulsion boundaries are $\varphi_1 = (x - 1)^2 + (y - 1.6)^2 - 0.01$ and $\varphi_2 = (x - 1)^2 + (y - 0.7)^2 - 0.04$. We selected \(n=4\) robots for the experiment. The experiment duration was 200 seconds, with a linear velocity \(v = 0.1m/s\), a maximum angular velocity \(\omega_{\max} = 3rad/s\), and a minimum turning radius \(r_{\min} = 0.033m\). It is noteworthy that in this experiment, the robots are treated as dynamic obstacles for collision avoidance among different robots.

The experimental results shown in Fig.~\ref{qwer} indicate that all robots are able to cooperatively move in the prescribed serial formation, successfully avoiding obstacles and other robots according to the designed obstacle avoidance strategy. Experimental video: https://www.bilibili.com/video/BV1EXbLzPE3X/.

\section{Conclusion}\label{wu}
This paper proposes a distributed safe cooperative vector field that effectively addresses the challenges of cooperative motion and safe collision avoidance in path-following tasks for trajectory curvature constrained multi-robot systems. By designing a safety-oriented collision avoidance vector field with adaptively adjustable reactive boundaries, the traditional fixed boundaries are replaced with real-time varying boundaries based on limit circles, thereby ensuring the physical feasibility of the collision avoidance behavior. The proposed vector field is composed of a cooperative motion vector field and a safety-oriented collision avoidance vector field, and relies solely on the virtual coordinates of neighboring robots to achieve cooperative motion, and ensure both obstacle avoidance and inter-robot collision avoidance. Experimental validation on a multi-robot platform demonstrates that the approach performs well in closed-path tracking, static obstacle avoidance, and inter-robot collision avoidance, while maintaining the effectiveness of cooperation. This approach provides a safety-guaranteed control solution for curvature-constrained multi-robot cooperation in complex environments.

%
%

\end{document}